\documentclass[11pt]{article}

\usepackage[final]{acl}
\usepackage{adjustbox}
\usepackage{multirow}
\definecolor{lightgray}{gray}{0.95}
\usepackage{booktabs}
\usepackage{colortbl}
\usepackage{xcolor}
\usepackage[ruled,vlined]{algorithm2e}
\usepackage{xspace}
\usepackage{amssymb}
\newcommand{\projectname}{\textsc{DocOCR-Eval}\xspace}

\usepackage{times}
\usepackage{latexsym}

\usepackage[T1]{fontenc}

\usepackage[utf8]{inputenc}

\usepackage{microtype}

\usepackage{inconsolata}

\usepackage{graphicx}

\title{\projectname: A Correction-Based Framework for OCR Tool Selection Without Ground Truth}

\author{
  \textbf{Zihan Xu\textsuperscript{1}},
  \textbf{Puzhen Wu\textsuperscript{2}},
  \textbf{Lawrence Chun Man Lau\textsuperscript{3}}, \\
  \textbf{Wei Liu\textsuperscript{4}}, 
  \textbf{Sirui Li\textsuperscript{5}},
  \textbf{Yifan Peng\textsuperscript{6}},
  \textbf{Yihao Ding\textsuperscript{4}}
\\
\\
  \textsuperscript{1}University of Melbourne,
  \textsuperscript{2}The University of Hong Kong, 
  \textsuperscript{3}The University of Hong Kong, \\
  \textsuperscript{4}University of Western Australia,
  \textsuperscript{5}Murdoch University,
  \textsuperscript{6}Weill Cornell Medicine
\\
  \small{
    \textbf{Correspondence:} \href{mailto:email@domain}{yihao.ding@uwa.edu.au}
  }
}

\begin{document}
\maketitle
\begin{abstract}
Document parsing is a foundational step for document understanding tasks such as visual question answering and key information extraction, as it transforms unstructured scanned images into structured representations by extracting textual, visual, and layout information. While numerous Optical Character Recognition (OCR) engines and multimodal large language models (MLLMs) have been developed for this purpose, selecting an appropriate document parsing solution for a given document collection remains challenging, particularly in label-scarce settings. In this work, we conduct a systematic evaluation of text recognition performance across a diverse set of OCR engines and state-of-the-art MLLMs on multiple scanned document benchmarks spanning different domains and languages. Motivated by the limited contextual reasoning capabilities of many OCR engines and the high cost of manual annotations, we propose \textbf{\projectname}, an annotation-free evaluation framework for automatic OCR assessment and selection. \projectname employs a three-staged correction and ranking strategy to approximate annotation-based tool ordering without ground-truth labels. We show that aggregating across multiple MLLMs progressively improves alignment with annotation-based rankings. Extensive experiments further demonstrate that reliable OCR tool selection can be achieved in realistic, label-limited settings, providing practical guidance for deploying document parsing systems across diverse real-world document collections.
\end{abstract}

\section{Introduction}

The need to understand visually rich scanned documents has arisen in many domains, including finance \cite{formnlu}, education \cite{vies}, and healthcare \cite{rxpad}. These documents are typically distributed as scanned images with complex layouts that combine text, visual elements, and structural cues, and are often affected by scanning noise and distortion. To support downstream tasks such as visual question answering \cite{docvqa,mmvqa} and key information extraction (KIE) \cite{sroie,ding2025synjac}, document parsing is required to convert unstructured images into structured representations by extracting textual, visual, and layout information.

%

\begin{table*}[ht]
\centering
\caption{Comparison of OCR engines, document parsing APIs, and vision-language models for document understanding.}
\footnotesize
\resizebox{\textwidth}{!}{%
\begin{tabular}{l l l l l l l l l}
\toprule
\textbf{Tool Name} & \textbf{Provider} & \textbf{Deployment} & \textbf{Pricing} & \textbf{Input Formats} & \textbf{Languages} & \textbf{Openness} & \textbf{GPU} & \textbf{Doc Parsing} \\
\midrule
\rowcolor{gray!20}\multicolumn{9}{l}{OCR Engine}\\
SuryaOCR~\cite{paruchuri2025surya} & Datalab & Local & Free & Image, PDF, Word, PPT & 90+ & Open & Optional & No \\
Kraken~\cite{kraken_ocr_2025} & Inria et al. & Local & Free & Image, PDF & Multi & Open & Optional & No \\
olmOCR~\cite{olmocrbench} & AI2 & Hybrid & Free & Image, PDF & Multi & Open & Required & Yes \\
AttentionOCR~\cite{Zhang2019-attention} & Guo \& Deng & Local & Free & Image & Multi & Open & Optional & No \\
Calamari~\cite{wick_calamari_2020} & Univ. Würzburg & Local & Free & Image & Multi & Open & Optional & No \\
EasyOCR~\cite{easyocr_2023} & JaidedAI & Local & Free & Image & 80+ & Open & Optional & No \\
Tesseract~\cite{tesseract_ocr_5} & S. Weil & Local & Free & Image & 100+ & Open & Not Required & No \\

PaddleOCR~\cite{cui2025paddleocr30technicalreport} & PaddlePaddle & Cloud & Free & Image, PDF & 80 & Open & Optional & Yes \\
docTR~\cite{doctr2021} & Mindee & Local & Free & Image, PDF & Multi & Open & Optional & No \\
Ocular~\cite{ocular_ocr} & Berkeley NLP & Local & Free & Image, PDF & Multi & Open & Optional & No \\
Google Cloud Vision~\cite{google_cloud_vision_ocr} & Google & Cloud & Paid & Image, PDF & 220+ & Closed & Not Required & Yes \\
 \midrule

\rowcolor{gray!20}\multicolumn{9}{l}{MLLM} \\
LightOnOCR~\cite{lightonocr2025}& LightOnAI & Cloud & Paid & Image, PDF & 2 & Closed & Required & No \\
Qwen3-VL~\cite{Qwen3-VL} & Aliyun & Hybrid & Free & Image, PDF & 32 & Closed & Required & No \\
OpenAI Vision~\cite{openai_vision_api} & OpenAI & Cloud & Paid & Image, PDF & Multi & Closed & Not Required & Yes \\
DeepSeek-OCR~\cite{wei2025deepseek} & DeepSeek AI & Hybrid & Paid & Image, PDF & Multi & Open-source & Supported & No \\
HunyuanOCR~\cite{hunyuanvisionteam2025hunyuanocrtechnicalreport} & Tencent & Local & Free & Image, PDF & 100+ & Open-source & Supported & No \\
Seed-VL~\cite{seed2025seed1_5vl} & ByteDance Seed & Cloud & Paid & Image, PDF & Multi & Open-source & Supported & Yes \\
\bottomrule
\end{tabular}
}
\label{tab:ocr_llm_comparison}
\end{table*}
OCR engines primarily focus on detecting and recognizing text from document images. They are typically open-sourced and freely available, but require non-image formats (e.g., PDF, Word, PowerPoint) to be converted into images prior to processing (Table~\ref{tab:ocr_llm_comparison}). While Optical Character Recognition (OCR) systems can extract text along with bounding boxes, often at the text-line level, they rely mostly on vision-based backbones trained in an object detection style. As a result, they are prone to missing text, spelling mistakes, and syntactic errors, which can propagate through OCR-dependent document understanding frameworks \cite{luo2024layoutllm, ding2025survey}. 

Recent multimodal large language models (MLLMs) offer an alternative by leveraging contextual reasoning for more robust text extraction, but they typically incur higher computational costs and rely on cloud-based deployment. Moreover, many MLLM-based methods generate plain-text outputs without explicit spatial boundaries, limiting their capability for layout-aware downstream systems and increasing the risk of hallucinations. 

Existing document understanding frameworks often rely on fixed or ad-hoc choices of parsing tools, overlooking the rapid evolution of both OCR engines and MLLM-based approaches. Moreover, evaluating text recognition performance typically requires costly manual annotations, which limit scalable and adaptive tool selection in real-world deployments. In practice, tool selection must balance recognition accuracy with resource constraints, such as annotation effort, computational overhead, and deployment cost.

To address this challenge, we propose an automated and cost-efficient framework for selecting the most suitable document-parsing tool for a given scanned document collection. The main contributions of this work are summarized as follows: 
\textbf{First}, we systematically evaluate conventional OCR engines and state-of-the-art MLLMs on representative scanned document benchmarks, providing a thorough cross-model and cross-domain comparison on text recognition.
\textbf{Second}, we propose \projectname, an annotation-free evaluation workflow that enables reliable cross-domain tool selection without requiring ground-truth labels.
\textbf{Finally}, we validate the effectiveness and robustness of \projectname across diverse document domains and languages, demonstrating its practical applicability in real-world settings.

\section{\textsc{DocOCR-Eval} Pipeline}
\subsection{Problem Definition} 
Given a document collection $\mathbb{D} = \{D_i\}_{i=1}^N$ from a specific domain and a set of $K$ candidate OCR engines $\mathcal{P} = \{p_j\}_{j=1}^K$, our goal is to automatically identify the engine $p \in P$ that yields the best overall performance on $\mathbb{D}$ without relying on human-annotated ground truth (GT). The selected OCR engine is then used to support downstream document parsing tasks, either by providing textual input to OCR-dependent frameworks or by serving as supervision for training OCR-free document understanding models.
\subsection{\projectname}
\begin{figure*}[t]
    \centering
    \includegraphics[width=0.8\linewidth]{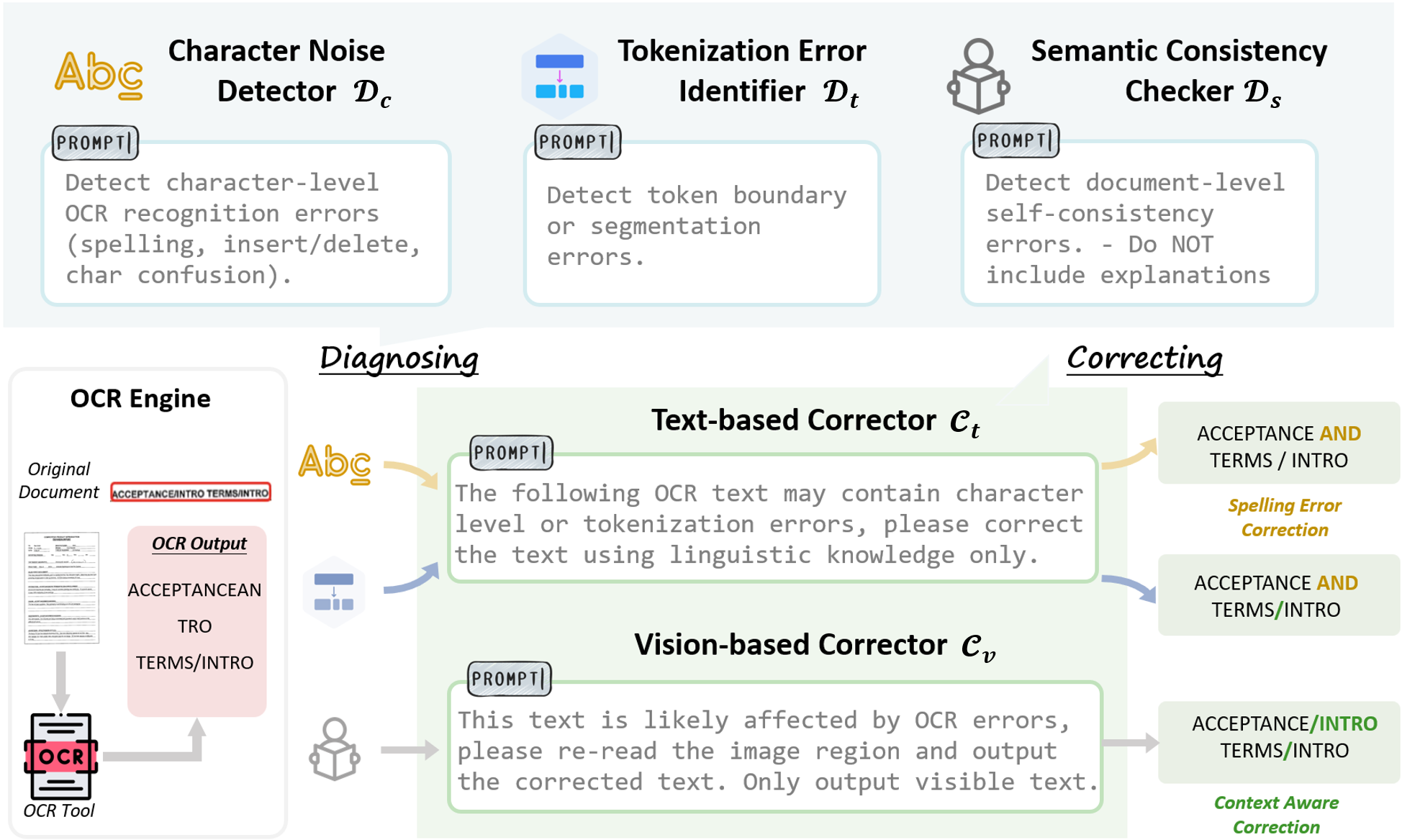}
    \caption{Workflow of the \projectname with a case study.}
    \label{fig:workflow}
\end{figure*}
We propose an annotation-free OCR engine selection pipeline that evaluates OCR quality by measuring the discrepancy between original OCR outputs and their corresponding corrections generated by an MLLM. 

For each document, the OCR output is processed through two stages: \emph{Error Diagnosing} and \emph{Error Correcting}.
By aggregating discrepancy scores over the entire document collection, we rank candidate OCR engines and select the one with the minimal overall discrepancy as the most suitable for deployment.

\subsubsection{Error Diagnosing Modules} 

These modules aim to diagnose various error types in the original OCR outputs (Figure~\ref{fig:workflow}). Given a document $D$, an OCR engine first extracts text and generates an output $T$. Each OCR output block $t \in T$ is then processed by three diagnosing modules. The Character Noise Detector ($\mathcal{D}_c$) detects character-level errors, including substitutions, insertions, deletions, and visually confusable glyphs. The Tokenization Error Identifier ($\mathcal{D}_t$) detects segmentation errors, such as incorrect word boundaries, whitespace errors, and line-break artifacts. The Semantic Consistency Checker ($\mathcal{D}_s$) assesses whether a text block is semantically inconsistent with the target document $D$. Each module outputs a binary decision, yielding a structured vector that enables fine-grained control over subsequent correction actions
$\mathbf{d}_k = [\mathcal{D}_c(t_k), \mathcal{D}_t(t_k), \mathcal{D}_s(t_k,D)]$.

\subsubsection{Error Correcting Modules}
We introduce two MLLM-based correction modules that leverage MLLM's rich commonsense knowledge and strong contextual reasoning.
The Text-based Corrector ($\mathcal{C}_t$) revises OCR outputs using linguistic and contextual cues alone, without introducing content beyond the recognized text. 
The Vision-based Corrector ($\mathcal{C}_v$) additionally incorporates the OCR bounding box cropped from the source image and instructs the MLLM to re-recognize only the visible text within that region.

\subsubsection{Error Diagnosis, Correction, and Ranking}

Based on the diagnostic signals from the error diagnosing modules ${\mathcal{D}_c,\mathcal{D}_t,\mathcal{D}_s}$, each OCR output block $t$ is conditionally corrected using the corresponding error correcting modules ${\mathcal{C}_t,\mathcal{C}_v}$.
Specifically, the Text-based Corrector $\mathcal{C}_t$ is applied when character-level or tokenization errors are detected, i.e., when $\mathcal{D}_c(t)=1$ or $\mathcal{D}_t(t)=1$. In contrast, the vision-based re-OCR corrector $\mathcal{C}_v$ is triggered for blocks exhibiting document-level semantic inconsistencies, indicated by $\mathcal{D}_s(t)=1$.

The staged correction yields post-correction outputs $\hat{T}_{D_i}^{(p_j)}$ for each OCR engine $p_j$ and document $D_i$. For single-MLLM correction, we compute Average Normalized Levenshtein Similarity (ANLS) scores $s_{j}$ for each engine and rank engines in descending order. To quantify the alignment between this correction-based ranking and the ground-truth ranking, we compute the Normalized Discounted Cumulative Gain (NDCG)~\cite{Wang2013-kb}, which assigns higher importance to top-ranked positions and progressively discounts lower-ranked items. Moreover, to mitigate potential bias from a single MLLM, we aggregate predictions from $k$ models by averaging their scores, $\bar{s}_{j}=\frac{1}{K}\sum_{k=1}^{K} s_{j}^{(k)}.$
\subsection{Experimental Setup}
\begin{table}[t]
\centering
\small
\setlength{\tabcolsep}{3pt}
\caption{Summary of adopted datasets. H: Handwritten. LA: Layout Analysis}
\label{tab:datasets_info}
\resizebox{\columnwidth}{!}{%
\begin{tabular}{l l l l r l}
\toprule
\textbf{Dataset} & \textbf{Language} & \textbf{Domain} & \textbf{H} & \textbf{Size} & \textbf{Proposed Tasks} \\
\midrule
FUNSD  & English   & Multiple              & Yes & 199  & OCR, LA, KIE \\
SROIE  & English   & Receipt               & No  & 1000 & OCR,  KIE \\
EPHOIE & Chinese   & Exam Paper & Yes & 1494 & OCR, KIE \\
RXPAD  & French    & Clinical Notes         & Yes & 200  & OCR, KIE \\
XFUND  & Multiple  & Multiple               & Yes & 1393 & OCR, LA, KIE \\
\bottomrule
\end{tabular}
}
\end{table}

\subsubsection{Dataset}
We evaluate OCR engines, MLLMs, and our OCR correction framework on five publicly available datasets covering diverse languages and domains: FUNSD~\cite{funsd} (general form understanding in English), SROIE~\cite{sroie}, EPHOIE~\cite{ephoie} (educational documents in Chinese), RXPAD~\cite{rxpad} (medical prescription documents in French), and XFUND~\cite{xu2022xfund} (Table~\ref{tab:datasets_info}). XFUND supports seven languages, each with 50 test images. To ensure fair and consistent evaluation across datasets of different sizes, we randomly sample 50 test images from each test split for all experiments.

\subsubsection{Implementation Details}

Table~\ref{tab:ocr_llm_comparison} summarizes the key attributes of each tool. Using annotated datasets, we directly compare the predictions of PaddleOCR~\cite{cui2025paddleocr30technicalreport}, EasyOCR~\cite{easyocr_2023}, Tesseract~\cite{tesseract_ocr_5}, Google Cloud Vision~\cite{google_cloud_vision_ocr}, Qwen3-VL-Instruct-4B~\cite{Qwen3-VL}, and GPT-4.1-mini~\cite{gpt4o} against ground truth across multiple datasets. These tools represent diverse OCR paradigms, including open-source rule-based engines (Tesseract), deep-learning OCR engines (PaddleOCR, EasyOCR), widely deployed commercial cloud services (Google Cloud Vision), and recent MLLMs with unified visual-text reasoning (Qwen3-VL-Instruct-4B and GPT-4.1-mini). 

For annotation-free evaluation, we evaluate PaddleOCR, EasyOCR, Tesseract, and Google Cloud Vision using MLLM-driven correction. For each OCR block, we apply a three-stage error diagnosis with Qwen3-VL-4B-Instruct, Qwen3-VL-8B-Instruct, GPT-4.1-mini, and Gemini-2.5-flash-lite, which triggers text- or vision-based re-OCR when needed. All experiments were conducted on NVIDIA A100 GPUs with deterministic inference by setting the temperature to 0.0 and disabling sampling.

\subsubsection{Evaluation Metrics}
For text recognition Performance, we report ROUGE-L to measure consistency between predicted and reference text, Character Error Rate (CER) to quantify character-level recognition errors, and ANLS as a length-normalised similarity measure. We also report word-level F1 to assess token-level coverage and the practical usability of extracted text. 

For tool selection performance, we assess ranking consistency using NDCG ($d$) under stepwise correction with each individual corrector ($\mathcal{D}_c$, $\mathcal{D}_t$, $\mathcal{D}_s$), as well as their cascaded combinations ($\mathcal{D}_c+\mathcal{D}_t$ and $\mathcal{D}_c+\mathcal{D}_t+\mathcal{D}_s$). Finally, an average NDCG score ($Avg.d$) is presented for each dataset. Performance is further evaluated quantitatively using ANLS for both single-MLLM and multi-MLLM post-correction settings. 
 
\section{Results and Discussion}

\subsection{OCR Tool Performance}
\begin{table}[t]
\centering
\caption{ANLS of text recognition with ground-truth.}
\label{tab:ocr_comparison}

\begin{adjustbox}{max width=\linewidth}
\begin{tabular}{lccccccccc}
\toprule
 & EPHOIE & FUNSD & SROIE & RXPAD & XF-ja & XF-es & XF-it & XF-de & XF-pt \\
\midrule
\rowcolor{gray!20}\multicolumn{10}{l}{OCR-engine-based}\\
PaddleOCR      & 0.20 & \textbf{0.42} & 0.51 & \textbf{0.66} & 0.46 & 0.83 & 0.34 & 0.88 & 0.88 \\
EasyOCR        & 0.09 & 0.29 & 0.35 & 0.61 & 0.53 & \textbf{0.87} & 0.74 & \textbf{0.90} & \textbf{0.90} \\
Tesseract      & 0.02 & 0.23 & 0.38 & 0.52 & 0.45 & 0.79 & 0.69 & 0.84 & 0.84 \\
Cloud Vision   & \textbf{0.44} & 0.29 & \textbf{0.56} & 0.29 & \textbf{0.62} & 0.73 & \textbf{0.75} & 0.74 & 0.74 \\
\midrule
\rowcolor{gray!20}\multicolumn{10}{l}{MLLM-based}\\
Qwen3-VL       & 0.46 & 0.29 & 0.53 & 0.28 & 0.48 & 0.59 & 0.46 & 0.70 & 0.70 \\
OpenAI         & 0.34 & 0.25 & 0.37 & 0.28 & 0.54 & 0.65 & 0.67 & 0.71 & 0.71 \\

\bottomrule
\end{tabular}
\end{adjustbox}
\vspace{-1em}
\end{table}


We first present text recognition results of representative OCR engines and MLLMs compared against ground-truth annotations. Table~\ref{tab:ocr_comparison} shows that no single OCR engine consistently outperforms the others, indicating that performance is highly dataset-dependent and motivating the need for automatic OCR tool selection. For instance, PaddleOCR performs strongly on FUNSD and RXPAD, while EasyOCR is competitive on several XFUND language subsets. By contrast, Cloud Vision and Qwen3-VL generally perform better on multilingual or visually complex datasets, such as EPHOIE and XFUND-it.

\begin{figure}[t]
  \centering
  \includegraphics[width=1.0\linewidth,clip,trim=0 4em 0 0]{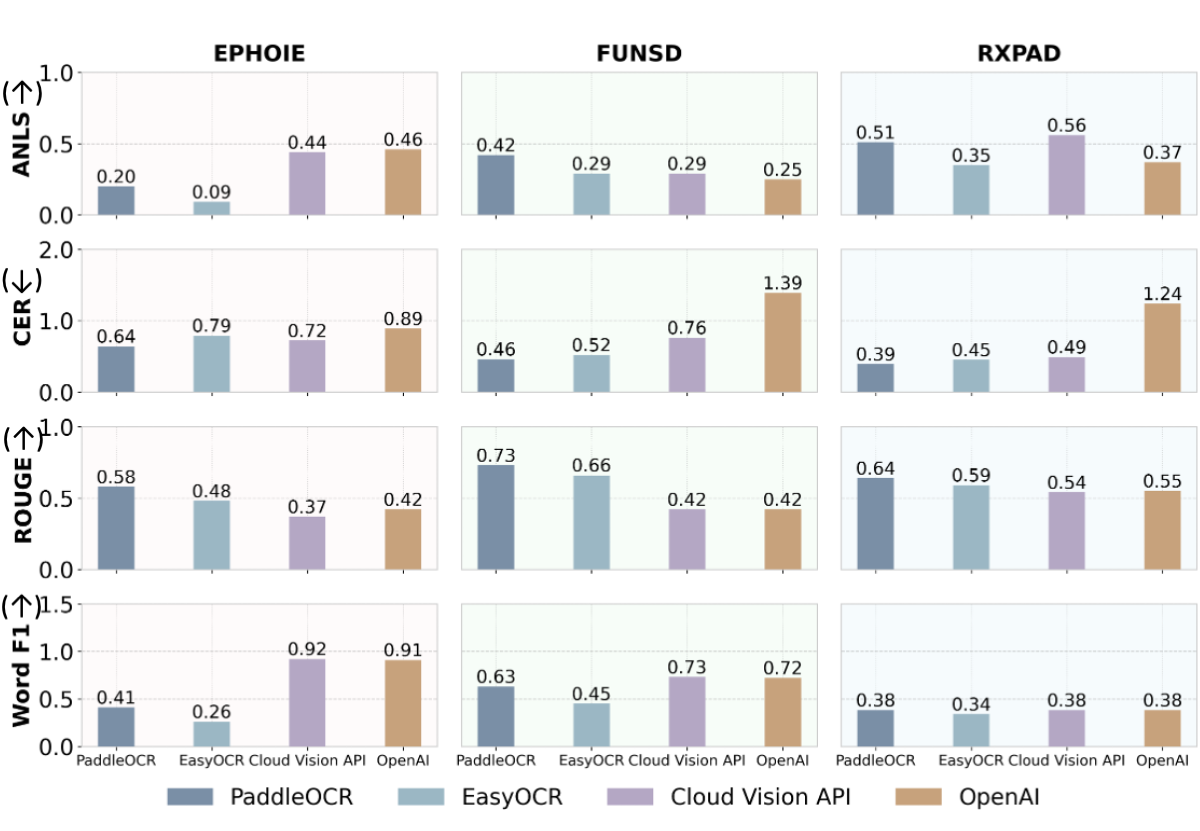}
  \caption{Metrics among diverse datasets by Qwen3-VL, $\uparrow$ higher is better; $\downarrow$ lower is better.}
  \label{fig:panel}
  \vspace{-1em}
\end{figure}

We then evaluate OCR results against ground truth across multiple metrics on EPHOIE, FUNSD, and RXPAD. Figure~\ref{fig:panel} shows that Cloud Vision API and OpenAI achieve higher ANLS and word-level F1, indicating stronger character- and word-level matching for non-English documents. In contrast, PaddleOCR often achieves lower CER and higher ROUGE, suggesting better lexical overlap and sequence-level fidelity.

\subsection{Performance of Automatic Selection}
\subsubsection{Best-Performing Tool Identification} 

Next, we evaluate the effectiveness of \projectname in automatically selecting the most suitable tool for domain-specific document collections. Table~\ref{tab:average_mllm} shows the average scores from four MLLMs. Compared with the single-MLLM correction (upper table), averaging ANLS across multiple MLLMs improves top-1 selection accuracy and narrows the performance gap between the annotation-based top-selected tool (highlighted in green) and competing alternatives. In the table, green cells denote the ground-truth best tool, bold values indicate the tool selected by correction, and red cells mark disagreements with the ground truth. Although the corrected rankings do not perfectly match the annotation-based rankings, the score differences consistently decrease as more models are aggregated. This trend indicates that averaging mitigates individual model bias and variance and produces rankings that increasingly approximate the ground-truth annotation-based order.
\begin{table}[t]
\centering
\caption{Text recognition results after correction. Green cells: the annotation-based top-1 tool. Bold: the tool selected for correction. Red cells: disagreements with GT.}
\label{tab:average_mllm}
\begin{adjustbox}{max width=\linewidth}
\begin{tabular}{lccccccccc}
\toprule
 & EPHOIE & FUNSD & SROIE & RXPAD & XF-ja & XF-es & XF-it & XF-de & XF-pt \\
 \midrule
\rowcolor{gray!20}\multicolumn{10}{l}{\textbf{Qwen3-VL post-correction ANLS}} \\
PaddleOCR      & 0.95 & \cellcolor{green!15}{\textbf{0.94}} & \cellcolor{red!15}{\textbf{0.95}} & \cellcolor{green!15}{\textbf{0.96}} & 0.92 & 0.97 & \cellcolor{red!15}{\textbf{0.98}} & \cellcolor{red!15}{\textbf{0.99}} & 0.95 \\
EasyOCR        & 0.47 & 0.82 & 0.90 & 0.92 & 0.87 & \cellcolor{green!15}{\textbf{0.98}} & 0.97 & \cellcolor{green!15}{0.97} & \cellcolor{green!15}{\textbf{0.96}} \\
Tesseract      & 0.56 & 0.64 & 0.93 & 0.88 & 0.58 & 0.92 & 0.93 & 0.96 & 0.91 \\
Cloud Vision   & \cellcolor{green!15}{\textbf{0.97}} & 0.88 & \cellcolor{green!15}{0.90} & 0.93 & \cellcolor{green!15}{\textbf{0.97}} & 0.95 & \cellcolor{green!15}{0.95} & 0.95 & 0.94 \\

\midrule
\rowcolor{gray!20}\multicolumn{10}{l}{\textbf{Average ANLS post-correction from multiple MLLMs}} \\
PaddleOCR  & 0.927 & \cellcolor{green!15}{\textbf{0.933}} & \cellcolor{red!15}{\textbf{0.945}} & \cellcolor{green!15}{\textbf{0.949}} & 0.767 & 0.941 & 0.946 & 0.949 & 0.910 \\
EasyOCR  & 0.467 & 0.805 & 0.890 & 0.888 & 0.747 & \cellcolor{green!15}{\textbf{0.953}} & 0.927 & \cellcolor{green!15}{\textbf{0.956}} & \cellcolor{green!15}{\textbf{0.939}} \\
Tesseract & 0.571 & 0.522 & 0.818 & 0.646 & 0.498 & 0.824 & 0.804 & 0.816 & 0.759 \\
Cloud Vision API & \cellcolor{green!15}{\textbf{0.949}} & 0.908 & \cellcolor{green!15}{0.912} & 0.928 & \cellcolor{green!15}{\textbf{0.940}} & 0.951 & \cellcolor{green!15}{\textbf{0.948}} & 0.951 & 0.935 \\
\bottomrule
\end{tabular}
\end{adjustbox}

\end{table}
\subsubsection{Ranking Consistency with ground truth} 

Table~\ref{tab:ranking_all_datasets} summarizes OCR engine rankings under annotation-based evaluation and our proposed stepwise correction framework, along with their rank alignment across datasets. Green cells indicate rankings consistent with the ground-truth order, while red cells denote discrepancies.
Overall, we observe moderate agreement between the two regimes under NDCG evaluation. On FUNSD ($Avg.d=0.9752$), correction-based rankings perfectly match annotation-based rankings, indicating that our framework can reliably recover relative performance without ground-truth labels. On more challenging datasets, alignment remains high but slightly reduced. EPHOIE achieves average NDCG values to 0.9679, while RXPAD reaches 0.9699, suggesting although exact rank matching may vary, the framework consistently preserves the most important top-ranked positions. 

\begin{table}[t]
\centering
\caption{OCR engine rankings under stepwise correction across datasets by Qwen3-VL. Green cells: rankings consistent
with the ground-truth order. Red cells: discrepancies in rankings with the ground-truth order.}
\footnotesize
\setlength{\tabcolsep}{3pt}
\label{tab:ranking_all_datasets}
\begin{adjustbox}{max width=\linewidth}
\begin{tabular}{
p{1.69cm}
>{\centering\arraybackslash}p{1cm}
>{\centering\arraybackslash}p{.95cm}
>{\centering\arraybackslash}p{.95cm}
>{\centering\arraybackslash}p{.95cm}
>{\centering\arraybackslash}p{.95cm}
>{\centering\arraybackslash}p{.95cm}
>{\centering\arraybackslash}p{.8cm}
}
\toprule
OCR Engine 
& GT Rank 
& $\mathcal{D}_c$
& $\mathcal{D}_t$
& $\mathcal{D}_s$
& $\mathcal{D}_c+\mathcal{D}_t$
& All 
& $Avg.d$ \\
\midrule

\rowcolor{gray!20}\multicolumn{8}{l}{EPHOIE}\\
PaddleOCR        & 2 & \cellcolor{red!15}{1} & \cellcolor{red!15}{1} & \cellcolor{green!15}{2} & \cellcolor{red!15}{1} & \cellcolor{green!15}{2} & \multirow{5}{*}{0.9679} \\
EasyOCR          & 3 & \cellcolor{green!15}{3} & \cellcolor{green!15}{3} & \cellcolor{green!15}{3} & \cellcolor{green!15}{3} & \cellcolor{red!15}{4} &  \\
Tesseract        & 4 & \cellcolor{green!15}{4} & \cellcolor{green!15}{4} & \cellcolor{green!15}{4} & \cellcolor{green!15}{4} & \cellcolor{red!15}{3} &  \\
Cloud Vision API & 1 & \cellcolor{red!15}{2} & \cellcolor{red!15}{2} & \cellcolor{green!15}{1} & \cellcolor{red!15}{2} & \cellcolor{green!15}{1} &  \\
$d$ & - & 0.9496 & 0.9496 & 1.0000 & 0.9496 & 0.9905 & \\
\midrule

\rowcolor{gray!20}\multicolumn{8}{l}{FUNSD}\\
PaddleOCR        & 1 & \cellcolor{green!15}{1} & \cellcolor{green!15}{1} & \cellcolor{red!15}{2} & \cellcolor{green!15}{1} & \cellcolor{green!15}{1} & \multirow{5}{*}{0.9752} \\
EasyOCR          & 3 & \cellcolor{red!15}{4} & \cellcolor{red!15}{4} & \cellcolor{green!15}{3} & \cellcolor{green!15}{3} & \cellcolor{green!15}{3} &  \\
Tesseract        & 4 & \cellcolor{red!15}{3} & \cellcolor{red!15}{3} & \cellcolor{green!15}{4} & \cellcolor{red!15}{2} & \cellcolor{green!15}{4} &  \\
Cloud Vision API & 2 & \cellcolor{green!15}{2} & \cellcolor{green!15}{2} & \cellcolor{red!15}{1} & \cellcolor{red!15}{4} & \cellcolor{green!15}{2} &  \\
$d$ & - & 0.9905 & 0.9905 & 0.9496 & 0.9543 & 1.0000 & \\
\midrule

\rowcolor{gray!20}\multicolumn{8}{l}{RXPAD}\\
PaddleOCR        & 1 & \cellcolor{red!15}{2} & \cellcolor{green!15}{1} & \cellcolor{green!15}{1} & \cellcolor{green!15}{1} & \cellcolor{green!15}{1} & \multirow{5}{*}{0.9699} \\
EasyOCR          & 2 & \cellcolor{red!15}{1} & \cellcolor{green!15}{2} & \cellcolor{red!15}{4} & \cellcolor{green!15}{2} & \cellcolor{red!15}{3} &  \\
Tesseract        & 3 & \cellcolor{green!15}{3} & \cellcolor{green!15}{3} & \cellcolor{green!15}{3} & \cellcolor{green!15}{3} & \cellcolor{red!15}{4} &  \\
Cloud Vision API & 4 & \cellcolor{green!15}{4} & \cellcolor{green!15}{4} & \cellcolor{red!15}{2} & \cellcolor{green!15}{4} & \cellcolor{red!15}{2} &  \\
$d$ & - & 0.9496 & 1.00 & 0.9543 & 1.00 & 0.9548 & \\
\bottomrule
\end{tabular}
\end{adjustbox}
\vspace{-2em}
\end{table}
We then examine the effect of each corrector. Table~\ref{tab:ranking_all_datasets} shows that optimal correction strategies vary across datasets. 
EPHOIE primarily benefits from context-aware re-reading (semantic correction $\mathcal{D}_s$), likely due to irregular layouts and semantic inconsistencies. 
FUNSD requires the full cascade, indicating heterogeneous error sources. 
In contrast, RXPAD favors text-only normalization ($\mathcal{D}_c + \mathcal{D}_t$), as domain-specific abbreviations make vision-based re-OCR prone to additional noise. 
These findings suggest that correction strategies should be dataset-adaptive rather than universally applied. 

\subsection{Case Study}

To illustrate how the proposed staged correction framework operates in practice, we present a representative case study drawn from the FUNSD dataset. Figure~\ref{fig:workflow} shows an OCR output produced by PaddleOCR and the subsequent correction process guided by our staged diagnostic pipeline using Qwen3-VL-Instruct-4B.

The original OCR output contains the phrase ``ACCEPTANCE-ANTRO TERMS/INTRO'', where the intended text is ``ACCEPTANCE /INTRO TERMS/INTRO''. Although the overall semantic structure is preserved, the text exhibits multiple subtle OCR errors.
In Step 1, the model identifies character-level errors such as missing accents and visually confusable characters (e.g., ``ACCEPRANCE-ANTRO'' vs. ``ACCEPTANCE''). 
In Step 2, it detects tokenization and segmentation errors, such as incorrect word boundaries and malformed token sequences (e.g., ``TERMS/INTRO'' instead of ``TERMS / INTRO''). These two steps operate independently and produce structured binary signals indicating whether each error type is present.
In Step 3, the model evaluates the OCR block against its surrounding page context. In this example, the phrase is semantically inconsistent with the surrounding text. As a result, vision-based re-OCR is triggered. Qwen3-VL correctly finds ``ACCEPTANCE/INTRO'', not ``ACCEPTANCE AND '', and returns the final corrected texts.

\subsection{Latency, Cost and Performance Trade Off}
While the proposed correction-based pipeline shows potential for annotation-free OCR tool selection, it introduces additional latency, because MLLMs are substantially slower than traditional OCR engines. Its usefulness therefore depends on downstream task requirements. For retrieval or coarse-grained indexing, lightweight OCR may suffice, whereas correction is better suited to high-accuracy settings. Moreover, commercial APIs and large vision–language models incur monetary or hardware costs, so users must balance performance gains against resource constraints. Accordingly, correction should be treated as a configurable option rather than a default solution.
\section{Conclusion}

This paper presents an annotation-free framework for OCR tool selection through a staged error detection and correction using MLLMs. By systematically comparing traditional OCR engines and MLLMs across multiple datasets, we demonstrate that correction-based evaluation can recover reliable tool rankings and identify top-performing systems without ground-truth annotations.

\bibliography{sigir2026}

@article{ding2025survey,
  title={A Survey on MLLM-based Visually Rich Document Understanding: Methods, Challenges, and Emerging Trends},
  author={Ding, Yihao and Luo, Siwen and Dai, Yue and Jiang, Yanbei and Li, Zechuan and Martin, Geoffrey and Peng, Yifan},
  journal={arXiv preprint arXiv:2507.09861},
  year={2025}
}

@article{ding2025synjac,
  title={SynJAC: Synthetic-data-driven Joint-granular Adaptation and Calibration for Domain Specific Scanned Document Key Information Extraction},
  author={Ding, Yihao and Han, Soyeon Caren and Li, Zechuan and Chung, Hyunsuk},
  journal={Information Fusion},
  pages={104074},
  year={2025},
  publisher={Elsevier}
}

@misc{paruchuri2025surya,
  author       = {Vikas Paruchuri and Datalab Team},
  title        = {Surya: A lightweight document OCR and analysis toolkit},
  year         = {2025},
  howpublished = {\url{https://github.com/VikParuchuri/surya}},
  note         = {GitHub repository},
}

@software{kraken_ocr_2025,
  title        = {The Kraken OCR system},
  author       = {Kiessling, Benjamin},
  year         = {2025},
  month        = aug,
  version      = {6.0},
  url          = {https://kraken.re},
  note         = {OCR system for historical and multilingual documents},
  howpublished = {\url{https://kraken.re}}
}

@misc{olmocrbench,
      title={{olmOCR: Unlocking Trillions of Tokens in PDFs with Vision Language Models}},
      author={Jake Poznanski and Jon Borchardt and Jason Dunkelberger and Regan Huff and Daniel Lin and Aman Rangapur and Christopher Wilhelm and Kyle Lo and Luca Soldaini},
      year={2025},
      eprint={2502.18443},
      archivePrefix={arXiv},
      primaryClass={cs.CL},
      url={https://arxiv.org/abs/2502.18443},
}

@ARTICLE{Zhang2019-attention,
  title         = "A feasible framework for arbitrary-shaped Scene Text
                   Recognition",
  author        = "Zhang, Jinjin and Wang, Wei and Huang, Di and Liu, Qingjie
                   and Wang, Yunhong",
  month         =  dec,
  year          =  2019,
  copyright     = "http://arxiv.org/licenses/nonexclusive-distrib/1.0/",
  archivePrefix = "arXiv",
  primaryClass  = "cs.CV",
  eprint        = "1912.04561"
}

@article{wick_calamari_2020,
    title = {Calamari - {A} {High}-{Performance} {Tensorflow}-based {Deep} {Learning} {Package} for {Optical} {Character} {Recognition}},
    volume = {14},
    number = {1},
    journal = {Digital Humanities Quarterly},
    author = {Wick, Christoph and Reul, Christian and Puppe, Frank},
    year = {2020},
}

@software{easyocr_2023,
  title        = {EasyOCR: Ready-to-use OCR with 80+ Supported Languages},
  author       = {JaidedAI},
  year         = {2023},
  url          = {https://github.com/JaidedAI/EasyOCR},
  note         = {Version referenced from GitHub repository},
  version      = {Version 1.7.2},
  howpublished = {\url{https://github.com/JaidedAI/EasyOCR}}
}

@software{tesseract_ocr_5,
  title        = {Tesseract OCR},
  author       = {{Tesseract OCR Project}},
  year         = {2023},
  version      = {5.0.0},
  url          = {https://tesseract-ocr.github.io/tessdoc/},
  note         = {Open source OCR engine},
  howpublished = {\url{https://tesseract-ocr.github.io/tessdoc/}}
}

@misc{cui2025paddleocr30technicalreport,
      title={PaddleOCR 3.0 Technical Report}, 
      author={Cheng Cui and Ting Sun and Manhui Lin and Tingquan Gao and Yubo Zhang and Jiaxuan Liu and Xueqing Wang and Zelun Zhang and Changda Zhou and Hongen Liu and Yue Zhang and Wenyu Lv and Kui Huang and Yichao Zhang and Jing Zhang and Jun Zhang and Yi Liu and Dianhai Yu and Yanjun Ma},
      year={2025},
      eprint={2507.05595},
      archivePrefix={arXiv},
      primaryClass={cs.CV},
      url={https://arxiv.org/abs/2507.05595}, 
}

@misc{doctr2021,
    title={docTR: Document Text Recognition},
    author={Mindee},
    year={2021},
    publisher = {GitHub},
    howpublished = {\url{https://github.com/mindee/doctr}}
}

@software{ocular_ocr,
  title        = {Ocular: Optical Character Understanding and Layout},
  author       = {{Ocular OCR Project}},
  year         = {2025},
  url          = {https://github.com/tberg12/ocular},
  note         = {Open source OCR tool for text recognition},
  howpublished = {\url{https://github.com/tberg12/ocular}}
}

@software{google_cloud_vision_ocr,
  title        = {Google Cloud Vision OCR},
  author       = {{Google LLC}},
  year         = {2025},
  url          = {https://docs.cloud.google.com/vision/docs/ocr},
  note         = {Cloud-based OCR service},
  howpublished = {\url{https://docs.cloud.google.com/vision/docs/ocr}}
}

@misc{lightonocr2025,
  title        = {LightOnOCR-1B: End-to-End and Efficient Domain-Specific Vision-Language Models for OCR},
  author       = {Said Taghadouini and Baptiste Aubertin and Adrien Cavaillès},
  year         = {2025},
  howpublished = {\url{https://huggingface.co/blog/lightonai/lightonocr}}
}

@article{Qwen3-VL,
      title={Qwen3-VL Technical Report}, 
      author={Shuai Bai and Yuxuan Cai and Ruizhe Chen and Keqin Chen and Xionghui Chen and Zesen Cheng and Lianghao Deng and Wei Ding and Chang Gao and Chunjiang Ge and Wenbin Ge and Zhifang Guo and Qidong Huang and Jie Huang and Fei Huang and Binyuan Hui and Shutong Jiang and Zhaohai Li and Mingsheng Li and Mei Li and Kaixin Li and Zicheng Lin and Junyang Lin and Xuejing Liu and Jiawei Liu and Chenglong Liu and Yang Liu and Dayiheng Liu and Shixuan Liu and Dunjie Lu and Ruilin Luo and Chenxu Lv and Rui Men and Lingchen Meng and Xuancheng Ren and Xingzhang Ren and Sibo Song and Yuchong Sun and Jun Tang and Jianhong Tu and Jianqiang Wan and Peng Wang and Pengfei Wang and Qiuyue Wang and Yuxuan Wang and Tianbao Xie and Yiheng Xu and Haiyang Xu and Jin Xu and Zhibo Yang and Mingkun Yang and Jianxin Yang and An Yang and Bowen Yu and Fei Zhang and Hang Zhang and Xi Zhang and Bo Zheng and Humen Zhong and Jingren Zhou and Fan Zhou and Jing Zhou and Yuanzhi Zhu and Ke Zhu},
	  journal={arXiv preprint arXiv:2511.21631},
      year={2025}
}

@software{openai_vision_api,
  title        = {OpenAI Images \& Vision API},
  author       = {{OpenAI}},
  year         = {2025},
  url          = {https://platform.openai.com/docs/guides/images-vision},
  note         = {Vision and image understanding API},
  howpublished = {\url{https://platform.openai.com/docs/guides/images-vision}}
}

@article{wei2025deepseek,
  title={DeepSeek-OCR: Contexts Optical Compression},
  author={Wei, Haoran and Sun, Yaofeng and Li, Yukun},
  journal={arXiv preprint arXiv:2510.18234},
  year={2025}
}

@misc{hunyuanvisionteam2025hunyuanocrtechnicalreport,
      title={HunyuanOCR Technical Report}, 
      author={Hunyuan Vision Team and Pengyuan Lyu and Xingyu Wan and Gengluo Li and Shangpin Peng and Weinong Wang and Liang Wu and Huawen Shen and Yu Zhou and Canhui Tang and Qi Yang and Qiming Peng and Bin Luo and Hower Yang and Xinsong Zhang and Jinnian Zhang and Houwen Peng and Hongming Yang and Senhao Xie and Longsha Zhou and Ge Pei and Binghong Wu and Kan Wu and Jieneng Yang and Bochao Wang and Kai Liu and Jianchen Zhu and Jie Jiang and Linus and Han Hu and Chengquan Zhang},
      year={2025},
      journal={arXiv preprint arXiv:2511.19575},
      url={https://arxiv.org/abs/2511.19575}, 
}

@article{seed2025seed1_5vl,
  title={Seed1.5-VL Technical Report},
  author={ByteDance Seed Team},
  journal={arXiv preprint arXiv:2505.07062},
  year={2025}
}

@inproceedings{mmvqa,
  title           = "{RotateKV}: Accurate and Robust 2-Bit {KV} Cache
                     Quantization for {LLMs} via {Outlier-Aware} Adaptive
                     Rotations",
  booktitle       = "Proceedings of the {Thirty-ThirdInternational} Joint
                     Conference on Artificial Intelligence",
  author          = "Su, Zunhai and Wei, Hanyu and Chen, Zhe and Shen, Wang and
                     Li, Linge and Yu, Huangqi and Yuan, Kehong",
  publisher       = "International Joint Conferences on Artificial Intelligence
                     Organization",
  pages           = "6200--6208",
  month           =  aug,
  year            =  2024,
  address         = "California",
  conference      = "Thirty-Third International Joint Conference on Artificial
                     Intelligence \{IJCAI-24\}",
  location        = "Jeju, South Korea"
}

@inproceedings{vies,
  title={Towards robust visual information extraction in real world: New dataset and novel solution},
  author={Wang, Jiapeng and Liu, Chongyu and Jin, Lianwen and Tang, Guozhi and Zhang, Jiaxin and Zhang, Shuaitao and Wang, Qianying and Wu, Yaqiang and Cai, Mingxiang},
  booktitle={Proceedings of the AAAI Conference on Artificial Intelligence},
  volume={35},
  number={4},
  pages={2738--2745},
  year={2021},
  url          = {https://doi.org/10.1609/aaai.v35i4.16378},
  publisher    = {{AAAI} Press},
}

@inproceedings{sroie,
  title           = "{ICDAR2019} competition on scanned receipt {OCR} and
                     information extraction",
  booktitle       = "2019 International Conference on Document Analysis and
                     Recognition ({ICDAR})",
  author          = "Huang, Zheng and Chen, Kai and He, Jianhua and Bai, Xiang
                     and Karatzas, Dimosthenis and Lu, Shijian and Jawahar, C V",
  publisher       = "IEEE",
  month           =  sep,
  year            =  2019,
  conference      = "2019 International Conference on Document Analysis and
                     Recognition (ICDAR)",
  location        = "Sydney, Australia"
}

@inproceedings{docvqa,
  title           = "{DocVQA}: A Dataset for {VQA} on Document Images",
  booktitle       = "2021 {IEEE} Winter Conference on Applications of Computer
                     Vision ({WACV})",
  author          = "Mathew, Minesh and Karatzas, Dimosthenis and Jawahar, C V",
  publisher       = "IEEE",
  month           =  jan,
  year            =  2021,
  copyright       = "https://ieeexplore.ieee.org/Xplorehelp/downloads/license-information/IEEE.html",
  conference      = "2021 IEEE Winter Conference on Applications of Computer
                     Vision (WACV)",
  location        = "Waikoloa, HI, USA"
}

@inproceedings{formnlu,
  title      = "{Form-NLU}: Dataset for the form natural language understanding",
  booktitle  = "Proceedings of the 46th International {ACM} {SIGIR} Conference
                on Research and Development in Information Retrieval",
  author     = "Ding, Yihao and Long, Siqu and Huang, Jiabin and Ren, Kaixuan
                and Luo, Xingxiang and Chung, Hyunsuk and Han, Soyeon Caren",
  publisher  = "ACM",
  pages      = "2807--2816",
  month      =  jul,
  year       =  2023,
  address    = "New York, NY, USA",
  copyright  = "https://www.acm.org/publications/policies/copyright\_policy\#Background",
  conference = "SIGIR '23: The 46th International ACM SIGIR Conference on
                Research and Development in Information Retrieval",
  location   = "Taipei Taiwan"
}

@inproceedings{funsd,
  title           = "{FUNSD}: A dataset for form understanding in noisy scanned
                     documents",
  booktitle       = "2019 International Conference on Document Analysis and
                     Recognition Workshops ({ICDARW})",
  author          = "Jaume, Guillaume and Kemal Ekenel, Hazim and Thiran,
                     Jean-Philippe",
  publisher       = "IEEE",
  month           =  sep,
  year            =  2019,
  copyright       = "https://ieeexplore.ieee.org/Xplorehelp/downloads/license-information/IEEE.html",
  conference      = "2019 International Conference on Document Analysis and
                     Recognition Workshops (ICDARW)",
  location        = "Sydney, Australia"
}

@inproceedings{luo2024layoutllm,
  title={LayoutLLM: Layout Instruction Tuning with Large Language Models for Document Understanding},
  author={Luo, Chuwei and Shen, Yufan and Zhu, Zhaoqing and Zheng, Qi and Yu, Zhi and Yao, Cong},
  booktitle    = {{IEEE/CVF} Conference on Computer Vision and Pattern Recognition,
                  {CVPR} 2024, Seattle, WA, USA, June 16-22, 2024},
  publisher       = "IEEE",
  volume          =  52,
  pages           = "15630--15640",
  month           =  jun,
  year            =  2024,
  conference      = "2024 IEEE/CVF Conference on Computer Vision and Pattern
                     Recognition (CVPR)",
  location        = "Seattle, WA, USA"
}

@misc{gpt4o,
  author = {{OpenAI}},
  title = {Hello GPT-4o},
  howpublished = {\url{https://openai.com/index/hello-gpt-4o/}},
  year = {2024},
}

@inproceedings{xu2022xfund,
  title={XFUND: A benchmark dataset for multilingual visually rich form understanding},
  author={Xu, Yiheng and Lv, Tengchao and Cui, Lei and Wang, Guoxin and Lu, Yijuan and Florencio, Dinei and Zhang, Cha and Wei, Furu},
  booktitle={Findings of the association for computational linguistics: ACL 2022},
  pages={3214--3224},
  year={2022}
}

@inproceedings{rxpad,
author = {Pattin Cottet, Jonathan and Eglin, V\'{e}ronique and Aussem, Alexandre},
title = {Rx-PAD: Recognition and\&nbsp;eXtraction – A\&nbsp;Dataset for\&nbsp;Prescription Analysis and\&nbsp;Clinical Data Structuring},
year = {2025},
isbn = {978-3-032-04629-1},
publisher = {Springer-Verlag},
address = {Berlin, Heidelberg},
url = {https://doi.org/10.1007/978-3-032-04630-7_9},
doi = {10.1007/978-3-032-04630-7_9},
booktitle = {Document Analysis and Recognition –  ICDAR 2025: 19th International Conference, Wuhan, China, September 16–21, 2025, Proceedings, Part V},
pages = {151–167},
numpages = {17},
location = {Wuhan, China}
}

@inproceedings{ephoie,
  title={Towards Robust Visual Information Extraction in Real World: New Dataset and Novel Solution},
  author={Wang, Jiapeng and Liu, Chongyu and Jin, Lianwen and Tang, Guozhi and Zhang, Jiaxin and Zhang, Shuaitao and Wang, Qianying and Wu, Yaqiang and Cai, Mingxiang},
  booktitle={Proceedings of the AAAI Conference on Artificial Intelligence},
  year={2021}
}

@ARTICLE{Wang2013-kb,
  title         = "A theoretical analysis of {NDCG} type ranking measures",
  author        = "Wang, Yining and Wang, Liwei and Li, Yuanzhi and He, Di and
                   Liu, Tie-Yan and Chen, Wei",
  month         =  apr,
  year          =  2013,
  copyright     = "http://arxiv.org/licenses/nonexclusive-distrib/1.0/",
  archivePrefix = "arXiv",
  primaryClass  = "cs.LG",
  eprint        = "1304.6480"
}




\end{document}